%% file: acl.tex
\pdfoutput=1

\documentclass[11pt]{article}

\usepackage{acl}

\usepackage{times}
\usepackage{latexsym}

\usepackage[T1]{fontenc}

\usepackage{tikz}

\newcommand{\udot}[1]{%
    \tikz[baseline=(todotted.base)]{
        \node[inner sep=1pt,outer sep=0pt] (todotted) {#1};
        \draw[dotted] (todotted.south west) -- (todotted.south east);
    }%
}%

\newcommand{\udensdot}[1]{%
    \tikz[baseline=(todotted.base)]{
        \node[inner sep=1pt,outer sep=0pt] (todotted) {#1};
        \draw[densely dotted] [->] (todotted.south west) -- (todotted.south east) ;
        \draw[densely dotted] [->] (todotted.south east) -- (todotted.south west) ;
    }%
}%

\newcommand{\uddot}[1]{%
    \tikz[baseline=(todotted.base)]{
        \node[inner sep=1pt,outer sep=0pt] (todotted) {#1};
        \draw[densely dotted] (todotted.south west) -- (todotted.south east) ;
        \draw[densely dotted] (todotted.south west) -- (todotted.south east) ;
    }%
}%

\newcommand{\dashdot}[1]{%
    \tikz[baseline=(todotted.base)]{
        \node[inner sep=1pt,outer sep=0pt] (todotted) {#1};
        \draw[thick,dash dot] (todotted.south west) -- (todotted.south east);
    }%
}%

\usepackage[utf8]{inputenc}

\usepackage{amsmath}
\DeclareMathOperator*{\argmax}{max}

\newcommand{\abs}[1]{\lvert#1\rvert}
\newcommand{\norm}[1]{\lVert#1\rVert}

\usepackage{amssymb}
\usepackage{listings}

\usepackage{booktabs}
\usepackage{graphicx}
\usepackage{multirow}

\usepackage{microtype}

\newcount\Comments  
\usepackage{color}
\definecolor{darkgreen}{rgb}{0,0.5,0}
\definecolor{purple}{rgb}{1,0,1}
\newcommand{\comment}[2]{\ifnum\Comments=1\textcolor{#1}{#2}\fi}

\usepackage{wasysym}

\usepackage{gb4e}
\noautomath

\title{Causal schema induction for knowledge discovery}

\begin{document}
\author{
    Michael Regan\textsuperscript{\dag\lightning},
    Jena D. Hwang\textsuperscript{\ddag},
    Keisuke Sakaguchi\textsuperscript{+},
    James Pustejovsky\textsuperscript{*} \\
    \textsuperscript{\dag} Paul G. Allen School of Computer Science \& Engineering, WA, USA \\
    \textsuperscript{\lightning} University of Colorado Boulder, CO, USA \\
    \textsuperscript{\ddag} Allen Institute for AI, WA, USA\\
    \textsuperscript{*} Brandeis University, MA, USA \\
    \textsuperscript{+} Tohoku University, Sendai, Japan \\
    \texttt{mregan@cs.washington.edu}
}

\maketitle

\begin{abstract}

Making sense of familiar yet new situations typically involves making generalizations about \textit{causal schemas},  stories that help humans reason about event sequences. Reasoning about events includes identifying cause and effect relations shared across event instances, a process we refer to as \textit{causal schema induction}. Statistical schema induction systems may leverage structural knowledge encoded in discourse or the causal graphs associated with event meaning, however resources to study such causal structure are few in number and limited in size.

In this work, we investigate how to apply schema induction models to the task of \textit{knowledge discovery} for enhanced search of English-language news texts. To tackle the problem of data scarcity, we present \textsc{Torquestra}, a manually curated dataset of text-graph-schema units integrating temporal, event, and causal structures. We benchmark our dataset on three knowledge discovery tasks, building and evaluating models for each. Results show that systems that harness causal structure are effective at identifying texts sharing similar causal meaning components rather than relying on lexical cues alone. We make our dataset and models available for research purposes.

\end{abstract}

\section{Introduction}
\label{sec:introduction}

Humans use language to understand stories describing participant interactions in events unfolding over time. To explain novel events in terms of previous experiences, humans rely heavily on \textit{causal schemas}:  stories about cause and effect relations that make memory and cognition more efficient \citep{Tversky1982,Kahneman2012}. If such schemas or stories form the basis of human reasoning, perhaps AI systems may similarly learn, store, and manipulate knowledge of causal structure for model interpretability or reasoning applications. However, datasets to support studies of causal schemas with natural language processing (NLP) methods are few and far between, a problem we set out to address in this work.

Making a dataset for the computational modeling of causal relations described in language is challenging, and so most existing resources are limited in size and focus on explicit causality at the sentence level. We introduce \textsc{Torquestra}, a dataset of implicit and explicit causal relations at the discourse level to support language studies using statistical methods (e.g., large language models). Our premise is that for human interpretability, causal stories are best represented as graphs, opening up decades of formidable research in graph theory that we can apply to our tasks.

\begin{figure}[t]
\scriptsize{
(\textbf{text})
{\fontfamily{qcs}\selectfont
   Karzai said he discussed the issue of \textbf{civilian casualties} when he held a \textbf{meeting} on \textbf{security} and \textbf{reconstruction}.}}

\begin{minipage}{.48\linewidth}
\centering
\captionsetup{justification=centering}
\includegraphics[width=15em]{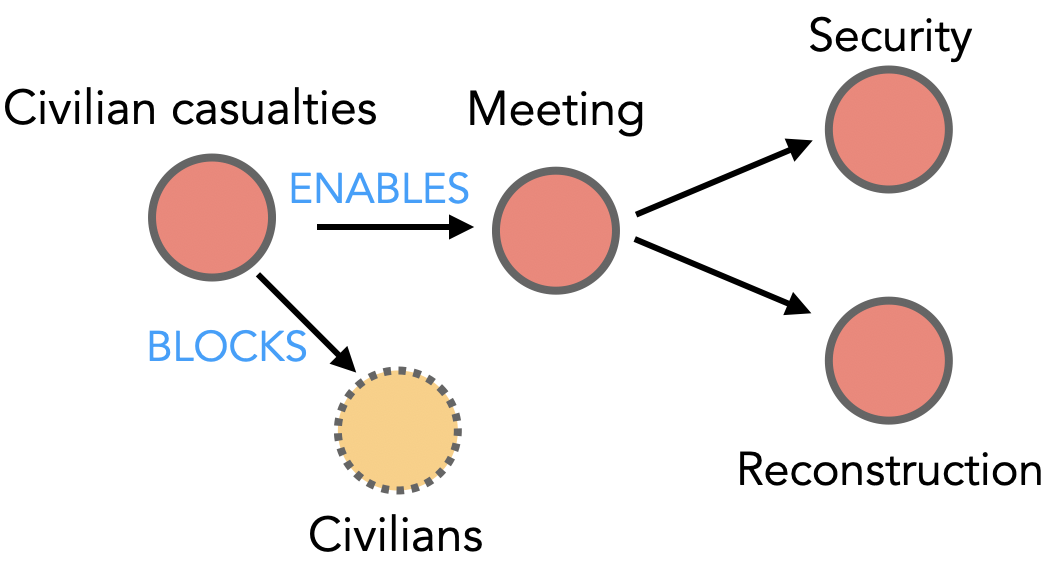}
  \caption*{\textsc{Causal instance graph}}
  
\end{minipage}
\hspace{0.02\linewidth}
\begin{minipage}{.48\linewidth}
\centering
\captionsetup{justification=centering}
\includegraphics[width=14em]{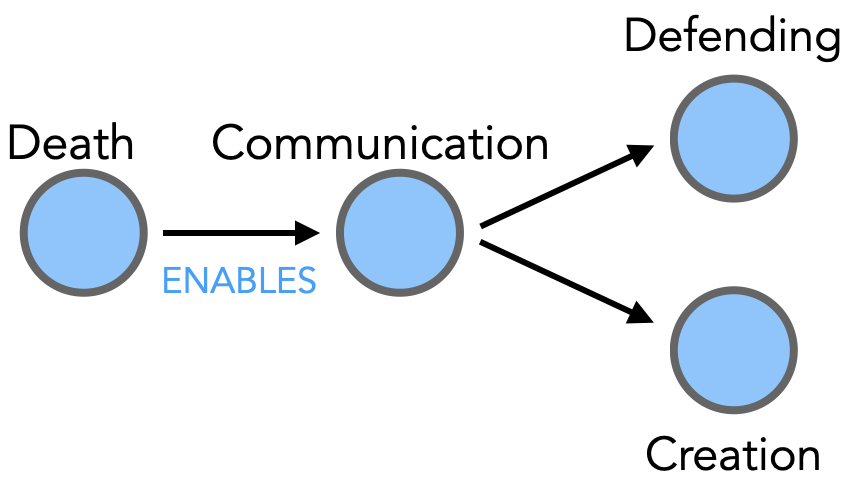}
  \caption*{\textsc{Causal schema graph}}
\end{minipage}

\caption{\label{fig:main}A causal schema is either an instance (left) tied directly to a text (top) or a schema graph (right) composed of event types. Edges indicate relations (not all shown) for causation of action and rest (\textsc{Enables} and \textsc{Blocks}). Graphs include participants, e.g., civilians (dotted orange node, left).}
\centering
\end{figure}

In Fig. \ref{fig:main}, we show a pair of \textsc{Torquestra} directed graphs. An instance graph (left) represents the causal story associated with a single text with short descriptions of events and participants as nodes. The corresponding schema (right) is a generalization of this causal story with event types for nodes, giving a means of inferring how different event instances may be similar in predictable ways.

Results of knowledge discovery experiments using \textsc{Torquestra} demonstrate that graph-based methods help identify texts that describe event sequences sharing similar causal structures as well as lexical features, with performance in clustering and schema matching experiments comparable to strong baselines that rely on lexical patterns alone. Through our experiments, we highlight the versatility of the dataset, with the hope of encouraging future research into causation and schemas in NLP.

As we study the inference of latent causal stories given textual descriptions of event sequences, our dataset, \textsc{Torquestra}, may help answer questions such as: (1) In what ways are temporal, causal, event hierarchical, and schema structures related? And, (2) How well do statistical methods such as pre-trained language models help with tasks that resemble causal reasoning? {To address these questions, our contributions include}:
\begin{itemize}\itemsep0em
\item  (\textbf{Theoretical}) We study participant-centered causal structure, a relatively unexplored approach to discourse modeling, for which we define fine-grained causal relations based on physical models of causation;
\item (\textbf{Dataset}) We present a dataset to analyze the temporal, event, schema, and causal structures described in natural language text; and,
\item (\textbf{Empirical}) We carry out experiments in structured generation for knowledge discovery, testing the suitability of a general purpose commonsense model distilled on symbolic knowledge for our data and tasks.

\end{itemize}

We first explore background in schema research (\S\ref{sec:schemas}) and define  causal structure from multiple perspectives including our own (\S\ref{sec:definitions}). We then take a close look at our dataset, \textsc{Torquestra}, including details about annotation and evaluation (\S\ref{sec:data}). 

To demonstrate the versatility of the dataset, our experiments include: causal instance graph generation, causal graph clustering, and causal schema matching, and we design and build models and metrics for each (\S\ref{sec:tasks}). We report baseline results with large language models and graph neural networks (\S\ref{sec:results}), concluding with remarks on challenges and opportunities of  schema understanding research.

\section{Causal schemas}
\label{sec:schemas}

Schemas, cf. \textit{scripts} and \textit{frames}, are high-level semantic structures for event sequences such as going to restaurants, crime investigations, and investing money  \citep{Minsky1974,Fillmore1976,Schank1977}, a coherent story or pattern of interactions distinct in memory. 

\subsection{Why schemas?}

Schemas help us reconstruct, order, and make predictions about events, about events' relative \textit{salience} and about the \textit{centrality} of event participants. Cognitive processes such as {generalization}, {induction}, and intuitive notions of physics and psychology \citep{Talmy1988,Tenenbaum2011} are associated with causal cues encoded in language \citep{Croft2012}. Together, these base elements give rise to \textit{causal reasoning}, a defining feature of human cognition and possibly one day of AI systems as well \citep{Lake2017,Scholkopf2021}.

\subsection{Causal schema induction}

In AI, semantic understanding or analysis is viewed as ``abduction to the best explanation'' \citep{Hobbs1993}. Abductive reasoning is tightly associated with \textit{induction}, which we view as abduction to the best \textit{high-level} explanation. The causal schema induction task is: given a text, infer high-level semantics for an event sequence using explicit (textual) and implicit (commonsense) knowledge. Consider the example events \ref{ex:salt1} and \ref{ex:money1}.

\begin{exe}
    \footnotesize
    \ex\label{ex:salt}  \begin{xlist}
        \ex\label{ex:salt1} I passed the salt to you.
        \ex\label{ex:salt2} \textsc{Transfer} $\nrightarrow$ \textsc{Change of ownership}
    \end{xlist}
    \ex\label{ex:money} \begin{xlist}
        \ex\label{ex:money1} I passed the money to you.
        \ex\label{ex:money2} \textsc{Transfer} $\rightarrow$ \textsc{Change of ownership}
    \end{xlist}    
\end{exe}

We formalize our notion of schemas using event types (\ref{ex:salt2} and \ref{ex:money2}) from FrameNet \citep{Fillmore2003}, with arrows denoting causal relations, either lack of enablement ($\nrightarrow$) or enablement ($\rightarrow$). This formalization helps represent  typical human experiences, e.g., not all \textsc{Transfer} events {enable} (or \textit{imply}, \textit{entail} or \textit{cause}) a \textsc{Change of Ownership}.

Human understanding of which logical conclusions are appropriate in a given context is integral to causal commonsense reasoning. In this paper, we examine how well large language models, e.g., GPT2/3 \cite{Radford2019,Brown2020}, perform related schema induction tasks.

\subsection{Related work}

In this section, we briefly describe relevant background in research on stories, event temporality, schemas, semantic search, and challenges of collecting causal data from natural language texts.

\textbf{Stories and time}. A \textit{story} is a temporal ordering of events \citep{Labov1967} characterized by change of state and participant interaction \citep{Croft2012,Croft2017}. In NLP, research into story understanding has emerged from studies of temporal relations \citep{Allen1983, Mani2006} using temporal data for model development and evaluation \citep{Pustejovsky2003}. Temporal event meaning is nuanced, evident in work on multiple meaning axes \citep{Ning2018b}, temporal aspect \citep{Donatelli2018}, and the relative duration of events \citep{Zhou2021}.

\textbf{Schemas as temporal structures}. Human knowledge is encoded in stories as \textit{schemas} \citep{Schank1995}, prototypical event sequences for common situations. In NLP, schemas are temporal structures, e.g., narrative event chains \citep{Chambers2008}, for event schema induction \citep{Chambers2013}, timeline construction \citep{Wen2021}, temporal schema induction \citep{Li2021}, future event prediction \citep{Li2022}, and partially-ordered temporal schema generation \citep{Sakaguchi2021}. However, temporal knowledge is complex and inherently noisy \citep{Ning2018b}, likely limiting advances in automated schema understanding systems.

\textbf{Temporal and causal structures are related}. Some lines of work integrate both temporal and causal perspectives, including narrative storylines \citep{Caselli2016} and representations for temporal and causal networks \citep{Bethard2008,Berant2014,Mirza2016,OGorman2018}. However much of this research does not directly address schemas, which we consider crucial for improved AI reasoning about stories, at the very least in an evaluation context.

\textbf{Temporal and causal datasets}. Our dataset is similar to efforts to crowdsource plot graphs \citep{Li2013}, collect graphical schemas for everyday activities \citep{Sakaguchi2021}, and apply text-graph pairs for temporal reasoning \citep{Madaan2021a}. Our work differs in that we integrate knowledge of causal, temporal, event, and schema structures in a single dataset.

\textbf{Semantic search}. Knowledge discovery can be framed as semantic search: identifying texts that share semantic structure. Heavy lifters in information retrieval are methods like BM25 and TF-IDF (sparse retrievers), often combined with text embedding similarity metrics (dense retrievers) \citep{Chen2022}. In work close to ours, similarity can also be measured using sentence meaning representations \citep{Bonial2020}, which we extend to study causal structure at the discourse level. 

\textbf{Challenges in making NLP causal datasets}. Due to the complexities of faithfully assessing causal relations, natural language datasets have focused mostly on explicit causal markers \citep{Mirza2014,Dunietz2017}  typically at the sentence level \citep{Tan2022}. In contrast, we seek to identify implicit (commonsense) and explicit causal relations at the discourse level, resulting in a dataset with at least 6x more causal relations than other resources, as we see in Table \ref{tab:datasets}.

\begin{table}[htbp]
    \scriptsize
    \centering
    \begin{tabular}{l|l|l}
\textbf{Dataset} & \textbf{\# docs} & \textbf{\# causal rels} \\
\hline
EventCausality \citep{Do2011}  & 25 & 580 \\
Causal-TB \citep{Mirza2014} & 183 & 318 \\
RED \citep{OGorman2016} & 90 & 1000 \\
Ning et al (\citeyear{Ning2018a}) & 25 & 172\\
ESTER \citep{Han2021} & 2000 & 4k \\
Causal News Corpus \citep{Tan2022}        & 3.5k sentences & {1600 (train)} \\
 Ours (\textsc{Torquestra$_{\textit{human}}$})       & 3k texts & 24k \\
 Ours (\textsc{Torquestra$_{\textit{auto}}$})       & 6k long texts & 75k \\
    \end{tabular}
    \caption{Existing human-curated datasets of causal relations in written text are relatively limited in size. Our base dataset (2nd from bottom) is at least 6x greater in number of causal relations compared to previous work.}
    \label{tab:datasets}
\end{table}

\section{Causal story framework}
\label{sec:definitions}

Causality is complex, as centuries of research stand to remind us. For one, commonsense causal reasoning goes beyond mere notions of necessary and sufficient conditions \citep{Minsky1974,Hobbs2005}. For another, causal viewpoints depend on perspective. Which causal dimensions of event sequences are humans most likely to agree on? To find answers to this question, we examine viewpoints from physics, neuroscience, philosophy, epidemiology, and cognitive semantics.

\subsection{Defining causal relations}
\label{ssec:relations}

\begin{table*}[t]
\scriptsize
\centering
\begin{tabular}{cllll}
\hline
\textbf{Rel} & \textbf{Sub-relation} & \textbf{Description} & \textbf{Verbs/concepts (exs.)} & \textbf{Example}\\
\hline
 \parbox[t]{2mm}{\multirow{6}{*}{\rotatebox[origin=c]{90}{\small{\textsc{Enables}}}}} &
 \textsc{Begins} & Prototypical causation of action & cause, start & Oleg started the ball rolling. \\
 &  \textsc{Adds} & Acceleration; cf. sufficient condition & contribute, help & Olga kept the ball rolling.\\
 &  \textsc{Allows/lets action} & Inaction allows action & let, allow, permit & Olaf let the ball roll (by not acting to stop it). \\
 &  \textsc{Prevents rest} & Remove barrier so action can continue & free, maintain & Oksana removed obstacles to the ball rolling.\\
 &  \textsc{Without effect} & Despite expectations, no enabling effect & despite, even though & Despite our efforts, we couldn't get the ball rolling. \\
 &  \textsc{Unknown} & Uncertainty of enabling relation & questions, modality & Did anybody/anything start the ball rolling?\\
\hline
  \parbox[t]{2mm}{\multirow{6}{*}{\rotatebox[origin=c]{90}{\small{\textsc{Blocks}}}}} &
 \textsc{Ends} & Prototypical causation of rest & stop & Oleg stopped the ball rolling. \\
 &  \textsc{Disrupts} & Reduction of momentum & hinder, resist, slow & Olga slowed the ball down. \\
 &  \textsc{Allows/lets rest} & Inaction leads to rest & not help & Olaf let the ball stop rolling. \\
 &  \textsc{Prevents action} & Barrier to action & refrain, forbid, hold & Oksana prevented the ball from rolling. \\
 & \textsc{Without effect} & Despite expectations, no blocking effect & despite, even though & We tried but could not stop the ball.\\
 &  \textsc{Unknown} & Uncertainty of blocking relation & questions, modality & Did anybody/anything stop the ball? \\   
\hline
\end{tabular}
\caption{\label{tab:relations}
Causal relations reflect the dual concepts of \textsc{Enables} ($\approx${makes more likely}), shorthand for \textit{causation of action}, and its counterpart \textsc{Blocks} ($\approx${makes less likely}), shorthand for \textit{causation of rest}. More fine-grained sub-relations (second column) are symmetric, e.g., the most prototypical causal relation  \textsc{Enables-Begins} corresponds to \textsc{Blocks-Ends}, \textsc{Enables-Adds} corresponds to \textsc{Blocks-Disrupts}, etc. The  sub-relation \textsc{Without effect} denotes the absence of expected causality for events that happen or do not happen \textit{despite expectations}, a challenging task for machines and often overlooked in other datasets.}
\end{table*}

In Newton's law of inertia \citep{Newton1687}, cause and effect relations are viewed in terms of \textbf{causation of action} (an external force puts an object into motion) and the \textbf{causation of rest} (an external force brings an object to rest). We argue that physical models for the acceleration and deceleration of an object through the addition of energy correspond closely with factors leading to the starting and ending of events, drawing on work in cognitive semantics \citep{Talmy1988,Croft2012} and psychology \citep{Wolff2007} where causal relations are conceived of as tendency to action and rest, force, opposition to force, and the overcoming of force.

Likewise, in neuroscience physical causal mechanisms are related to excitatory and inhibitory synapses  making neuron responses either more or less likely \citep{Purves2017}, similar to views in scientific philosophy \citep{Reichenbach1956} and causation in epidemiology \citep{Parascandola2001}. Across views, causal factors are associated with that which increases or decreases the likelihood of events, notions we integrate into  definitions for causal structure we present in Table \ref{tab:relations}.

\subsection{Participant-centered causal structure} 

Modeling causal relations depends not only on how one event (a precondition or \textit{cause}) is related to a subsequent event (a postcondition or \textit{effect}), but also to the direct role of event \textit{participants}, often the grammatical subjects and objects associated with event structure \citep{Talmy1988,Croft2012}. Participant-centeredness is featured in studies of narrative \citep{Propp1968,Caselli2016,Brahman2021}, of participant states \citep{Ghosh2022,Vallurupalli2022}, and of disease where organisms are conceived of as causative agents, e.g., the pathogen \textit{tubercle bacillus} \textsc{Causes} \textit{tuberculosis}.

\begin{figure}[htbp]
\centering
\includegraphics[trim={0 0.2cm 0 0.3cm},clip,width=0.22\textwidth]{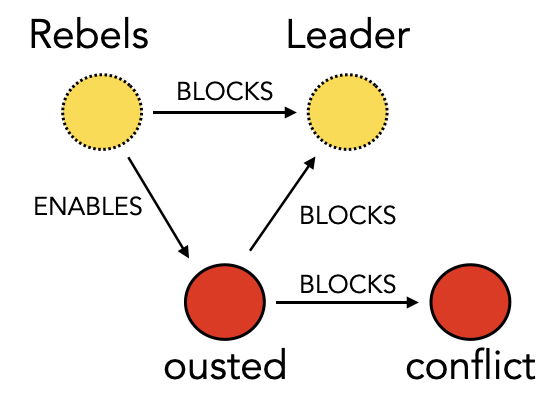}
\caption{\label{fig:instance-graph} An example participant-centered instance graph for ``The rebels ousted the leader to end the conflict.'' Causal graphs consist of two types of nodes: participants (top, orange) and events (bottom, red). Here, the rebels enable the ousting event, thus blocking the leader and the conflict as well.}
\end{figure}

In the participant-centered graph in Fig. \ref{fig:instance-graph} we show how participants and events causally interact. Typical causal relations are between events, e.g., the \textit{ousting of a leader} may end or \textsc{Block} a \textit{conflict}. In a participant-centered approach, people and things directly act on one another and also act as the initiating agents or causal endpoints of events, e.g., the \textit{rebels} \textsc{Block} the \textit{leader} who in turn \textsc{Enables} the conflict, etc.

\section{\textsc{Torquestra}}
\label{sec:data}

\textsc{Torquestra} is a \textit{causal schema library}: a dataset\footnote{\url{https://github.com/fd-semantics/causal-schema-public}} of text-graph-schema units for research in schema induction and, more broadly, knowledge discovery (see \S\ref{sec:tasks}). At its core, a \textsc{Torquestra} data instance is a newswire text paired with causal ($G_{\textit{causal}}$), temporal ($G_{\textit{temp}}$) and event ($G_{\textit{event}}$) structures. For notation, see Table \ref{tab:symbols}.

\begin{table}[htbp]
\centering
\small
\begin{tabular}{cl}
\hline
\textbf{Symbol} & \textbf{Meaning} \\
\hline
$G_{\textit{causal}} \in \mathcal{G}$ & Instance causal graph assoc. w/ text\\
$S_{\textit{causal}} \in \mathcal{S}$ & Schema causal graph assoc. w/ text(s)\\
$p \in \mathcal{P}$ & Participant node in a causal graph\\
$\langle p_i, rel, p_j \rangle$ & Causal relation (see Table \ref{tab:relations}) \\
{} &  between participants $p_i$ and $p_j$ \\
$G_{\textit{temp}}$ & Temporal graph \citep{Ning2020}\\
$G_{\textit{event}}$ & Event graph \citep{Han2021}\\
$V_{\textit{maven}}$ & Node w/ event types \citep{Wang2020}\\
$\phi_{\textit{maven}}$ & Set of hierarchical event types \\
\hline
\end{tabular}
\caption{\label{tab:symbols}
Notation used in this paper.
}
\end{table}

In this section, we briefly describe the  resources we used to make \textsc{Torquestra}, including notes about texts, size, graphs, and data annotation.

\subsection{Temporal and event structure knowledge} 

\begin{figure}[htbp]
\centering
\includegraphics[width=0.40\textwidth]{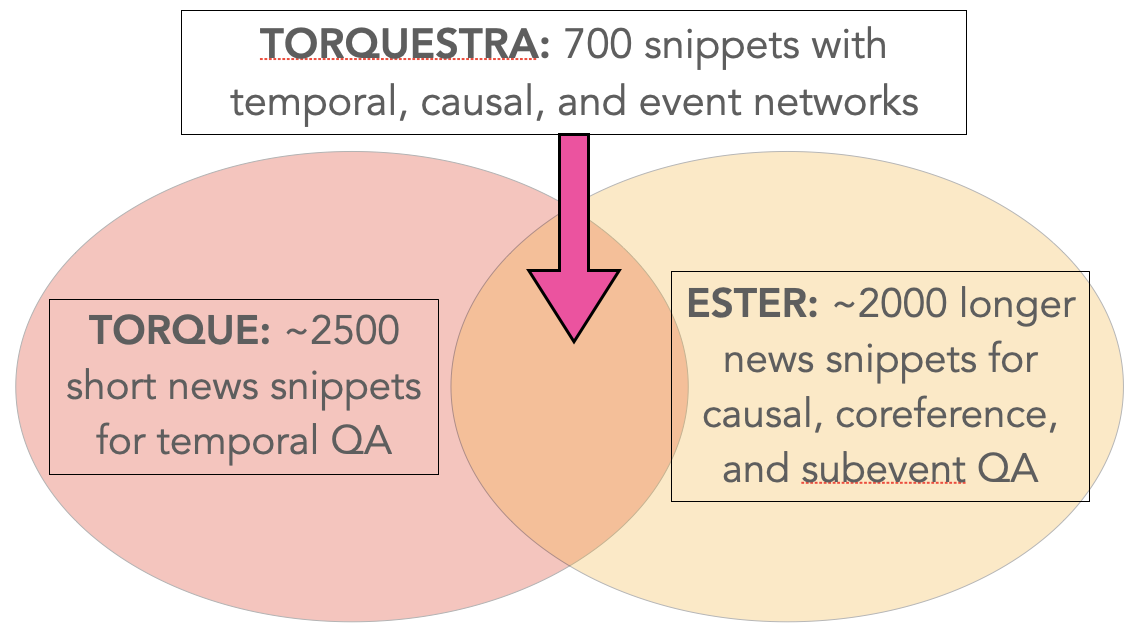}
\caption{\label{fig:torquestra}The core of \textsc{Torquestra} is drawn from texts with rich QA annotations from two existing resources: \textsc{Torque} \citep{Ning2020} and \textsc{Ester} \citep{Han2021}.}
\end{figure}

\textbf{\textsc{Torque} +  \textsc{Ester} =  \textsc{Torquestra}}. We wanted a dataset for a joint study of temporal, causal, and event structures. To this end, we examined question-answer (QA) datasets for temporal relations \textsc{Torque} \citep{Ning2020} and event structures \textsc{Ester} \citep{Han2021}. After noting both drew texts from TempEval3, a subsequent analysis revealed the datasets shared 700 text snippets, an intersection of data that we used to form core \textsc{Torquestra}, as illustrated in Fig. \ref{fig:torquestra}.

\subsection{Texts}

\textbf{Texts} in \textsc{Torquestra} are English-language newswire snippets from TempEval3 \citep{Uzzaman2013} and Wikipedia. The texts cover a number of typical news domains, including politics, sports, and business. Texts are mostly multiple sentences (98\%+), with mean text length between 60-300 subword tokens (Byte Pair Encoding \citep{Sennrich2016}).

\subsection{Dataset size}

Our manually constructed dataset consists of three slices of data: \textsc{Torque} (2500 exs), \textsc{Ester} (700 exs), and \textsc{Wiki-crime} (200 exs), each  aligned with up to four semantic networks. For details about data slices, see Appendix \ref{ssec:describing-torquestra}, Fig. \ref{fig:torquestra-content}.

\subsection{Causal instance graphs}
\label{ssec:instance-graphs}

\textsc{Torquestra} consists of causal instance graphs for events described in text (see Fig. \ref{fig:instance-graph}). A causal graph $G_{\textit{causal}}=(V,E)$ is directed and at times cyclic,  with vertices $V$ for salient events and participants and edges $E$ being causal relations. Nodes in the graphs are natural language descriptions for events and event participants, typically of subject-verb-object form. For graph statistics, see Appendix \ref{ssec:describing-torquestra}, Fig. \ref{fig:torquestra-graphs}.

\subsection{Causal schema graphs}
\label{ssec:schema-graphs}

As counterparts to $G_{\textit{causal}}$, schema causal graphs $S_{\textit{causal}}$ are generalizations for event sequences using event types from $V_{\textit{maven}} \subset \phi_{\textit{maven}}$\footnote{ $|\phi_{\textit{maven}}|=168$ FrameNet event types}. Annotators also add free-form event labels for cases not represented in $\phi_{\textit{maven}}$, e.g., event types such as \textsc{International\_relations}, amounting to >$400$ event types observed (with details in Appendix \ref{ssec:appendix-events}).

\begin{figure}[htbp]
\centering
\includegraphics[width=0.35\textwidth]{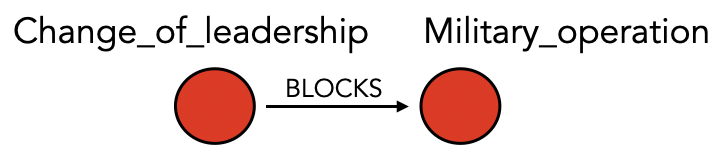}
\caption{\label{fig:schema-graph} A causal schema graph for the event sequence ``The ousting of the leader ended the conflict'' using frame semantic labels as nodes.}
\end{figure}

Compare the instance graph from earlier (Fig. \ref{fig:instance-graph}) to the schema graph in Fig. \ref{fig:schema-graph}: the `ousting of the leader' blocks the `conflict' in the instance graph, which in the schema graph is generalized to a 
\textsc{Change of leadership} that blocks a \textsc{Military operation}.

Ontological questions for causal schemas remain open. In \textsc{Torquestra}, causal schema are associated with event frame semantics at the node- or subgraph-level, one step of the data collection process that we discuss next.

\subsection{Data annotation}
\label{ssec:annotation}

The dataset is compiled through manual and automated means, subsets of data we refer to as \textsc{Torquestra}$_{\textit{human}}$ and \textsc{Torquestra}$_{\textit{auto}}$. In this subsection, we focus on manual annotation efforts.

\textbf{\textsc{Torquestra}$_{\textit{human}}$} consists of approximately 30K spans of text corresponding to graph nodes and 48K additional labels for nodes and edges. Annotation consisted of four main tasks: Given a short text and commonsense knowledge, identify and label causal participants (nodes), event types (for nodes/graph), causal relations (edges), and salient causal chains.

For annotation, we relied on a group of eight (8) in-house undergraduate and graduate students with backgrounds in linguistics and computer science which we found could faithfully recreate the causal graphs we envisioned. Core \textsc{Torquestra}$_{\textit{human}}$ required approx 250 hours with annotators earning between \$16-25/hr. For more details about the annotation process, guidelines, evaluation, and prompt engineering, see Appendix \ref{ssec:appendix-annotation}.

\section{Tasks and experimental methods}
\label{sec:tasks}

\begin{figure}[htbp]
\centering
\includegraphics[width=0.48\textwidth]{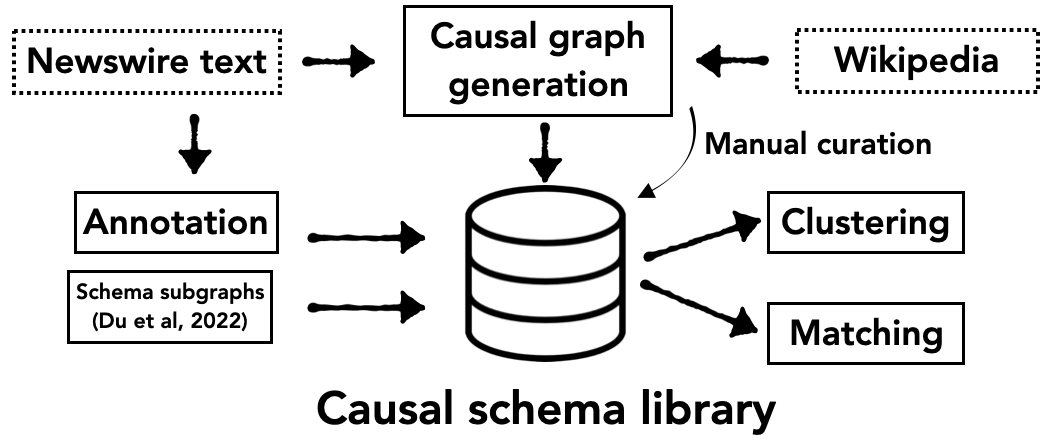}
\caption{\label{fig:library} Data-task pipeline. With texts from different sources (dotted boxes), we annotate, generate, and collect causal stories for a \textit{schema library}, a repository we use for \textit{causal graph clustering} and \textit{schema matching}.}
\end{figure}

\textsc{Torquestra} supports the induction and knowledge discovery tasks illustrated in Fig. \ref{fig:library}: (1) \textit{causal instance graph generation}, (2) unsupervised  
\textit{causal graph clustering}, and (3) \textit{causal schema matching}. We briefly describe each in turn.

\subsection{Causal instance graph generation}
\label{ssec:generation}

The first task is \textit{causal instance graph generation}. As in work on narrative planning \cite{Riedl2010}, learning mini knowledge graphs as world models \citep{Ammanabrolu2021}, and temporal graph generation \citep{Madaan2021a}, we generate graphs conditioned on text and, in an extension to previous work, also condition on event semantics (e.g., temporal structures). We compare few-shot GPT-3 \citep{Brown2020} with fine-tuned GPT2-XL and knowledge distilled GPT2-XL$_{\textit{distill}} $\citep{West2022} (60/40 train/dev split).

\textbf{Generation evaluation}. Perplexity and n-gram overlap metrics such as METEOR \citep{Banerjee2005} are of limited use as proxy measures for the faithfulness and coherence of generated causal stories. So, we also manually evaluate triples, reporting \textit{correctness} (\% of accurate edges) and \textit{completeness} (\% of causal graph generated).

\begin{table*}[t]
\centering
\footnotesize
\begin{tabular}{lllcccc}
\toprule
\textbf{Experiment} & \textbf{Model} & \textbf{Output | Input} &  \multicolumn{4}{c}{\textbf{Metrics}}\\
{} & {} & {} & \scriptsize{METEOR} & \scriptsize{Correct} & 
\scriptsize{Complete} & \scriptsize{\# triples eval} \\ 
\toprule

{Graph generation 1} & GPT-3 (7-shot) & \scriptsize{$p(G_{\textit{causal}} | \textit{text},G_{\textit{temp}})$}  & {0.28}  & \udensdot{0.50} & {0.33} & {224} \\

{(validation set)} & {} & \scriptsize{$p(G_{\textit{causal}} | \textit{text},G_{\textit{event}})$}  & {0.26} &  {0.55} & {0.25} & {180} \\

{(supervised, 60/40 split)} & GPT2-XL & \scriptsize{$p(G_{\textit{causal}} | \textit{text},G_{\textit{temp}})$}  & {0.27}  & \udot{0.52} & \dashdot{0.23} & {180} \\

{} & GPT2-XL$_{\textit{distill}}$ & \scriptsize{$p(G_{\textit{causal}} | \textit{text})$}  & {0.27}  & \underline{0.58} & \underline{\underline{0.35}} & {120}\\

{} & {} & \scriptsize{$p(G_{\textit{causal}} | \textit{text},G_{\textit{temp}})$}  & {0.29}  & \udensdot{0.60} & \dashdot{0.38} & {120}\\

{} & {} & \scriptsize{$p(G_{\textit{causal}} | \textit{text},V_{\textit{event}})$}  & {0.34}& {0.60} & {0.40} & {300}\\

{} & {} & \scriptsize{$p(G_{\textit{causal}} | \textit{text},G_{\textit{temp}},V_{\textit{event}})$}   & \textbf{0.41} & \underline{\textbf{0.65}} & \underline{\underline{\textbf{0.42}}} & {360}\\

\midrule

{Graph generation 2} & GPT2-XL$_{\textit{distill}}$  &  \scriptsize{$p(G_{\textit{causal}} | \textit{text},G_{\textit{temp}})$}  & {n/a}  & {0.56} & {0.33}  & {320} \\

{(test set)} & {} &  \scriptsize{$p(G_{\textit{causal}} | \textit{text}, V_{\textit{event}})$}  & {n/a}  & {0.59} & {0.36} & {360}\\

\bottomrule
\end{tabular}
\caption{\label{tab:results}
Results for causal graph (G$_{\textit{causal}}$) generation using GPT-3 \citep{Brown2020}, GPT2-XL \citep{Radford2019} and GPT2-XL$_{\textit{distill}}$ \citep{West2022}. We condition on texts + temporal networks ($G_{\textit{temp}}$), event structure ($G_{\textit{event}}$), and hierarchical events ($V_{\textit{event}}$), automatically evaluating with METEOR and manually evaluating correctness and completeness. Pairs of results \underline{underlined} (or dotted) illustrate important points we discuss in \S\ref{sec:results}.
 }
\end{table*}

\subsection{Causal graph clustering}

The second task is \textit{causal graph clustering}, unsupervised clustering of schema instance graphs (\textsc{Torquestra$_{\textit{auto}}$}). The objective of this task is to study the effectiveness of similarity metrics for texts using lexical features and graph embeddings.

\textbf{Data}. As out-of-domain data for testing, we used MAVEN (MAssive eVENt detection) \citep{Wang2020}, a collection of 3K+ Wikipedia articles that targets open domain event understanding systems, adopting 168 labels from FrameNet \citep{Fillmore2003} organized in an event hierarchy. 

\textbf{Models}. For lexical similarity baselines, we use standard implementations of tf-idf\footnote{\footnotesize{https://en.wikipedia.org/wiki/Tf-idf}} for sparse vectors and a SentenceTransformer \citep{Reimers2019} for dense text embeddings. 

For graph embeddings, we first encode graph nodes using DeBERTa (900M model) \citep{He2021}, and assign scalar values to edges ($+1$ for \textsc{Enables} and $-1$ for \textsc{Blocks}). We then train a graph attention network  \citep{Velickovic2018} via self-supervision masking random nodes. For further comparison, we also compose graph embeddings using the FEATHER algorithm \citep{Rozemberczki2020} based on random walks.

We cluster embeddings using standard {K-means}\footnote{\footnotesize{https://scikit-learn.org/stable/modules/clustering.html}} with $k$=6 for the number of clusters and fix the number of observations to $n$=25 for evaluation. We then examine results, measuring similarity using automated metrics we outline next.

\textbf{Clustering metrics}. We report purity, adjusted Rand index (ARI), and V-measure (VM) \citep{Rosenberg2007} using as ground truth topic labels for each article \citep{Wang2020} mapped to a smaller ontology\footnote{e.g., `hurricane' and `earthquake' are both \textsc{Disaster}}. 

To measure similarity accounting for multiple labels, we propose a new metric: \textit{event cluster purity}, estimating for each cluster a true label $e_j$ as the top-j event types observed\footnote{\footnotesize{Reminiscent of Jaccard similarity}}. In these experiments, we look at the top-10 event types observed, $j$=10, so $e_{10}^m$ denotes the ground truth event type vector for cluster $m$. We then compare this ground truth with a human-annotated k-hot event vector for each graph, $e^G$. For $N$ clusters $M$, the metric is defined:\begin{equation}
\label{eq:purity}
\text{purity}_{\textit{event}}=\frac{1}{N}\sum_{m\in M}  \sum_{e^G\in m} \argmax_{e_{10}^m \in \textbf{E}}|e_{10}^m \cap e^G| 
\end{equation}

\subsection{Causal schema matching}

As a variant of exemplar matching, the aim of causal schema matching is to identify induced MAVEN graphs (\textsc{Torquestra$_{\textit{auto}}$}) most similar to  curated schemas from two sources: core \textsc{Torquestra$_{\textit{human}}$} (short texts) and an existing schema library \citep{Du2022}. In the latter case, we match to schema chapters (individual subgraphs) for ease of evaluation. 

Previous work has investigated methods to align schema nodes \citep{Du2022}, and   methods for subgraph matching exist \citep{Rex2020}. Nonetheless, our experiments show the effectiveness of using graph embedding similarity as a step in identifying relevant schemas given a query.

Using the same models as our clustering experiments, we randomly select 50 Wikipedia articles to match with our schema library (RESIN \citep{Du2022} + core \textsc{Torquestra}) and examine the top-5 matched schemas using text topic labels, event type overlap, and graph visualizations for qualitative analysis. We report mean average precision (MAP) and mean reciprocal rank (MRR) as the accuracy of the ranking of the most relevant text. For more details about metrics and the evaluation tool, consult the Appendix \ref{ssec:metrics} and website.

\section{Results and discussion}
\label{sec:results}

We present results for causal instance graph generation using manual and automatic evaluation in Table \ref{tab:results} and results for causal graph clustering and schema matching in Tables \ref{tab:results-clustering} and \ref{tab:results-matching}.

We report mean results for a minimum of three different model runs varying random seeds (of graph neural networks) and hyperparameters (\# epochs, block size, p-sampling rate, etc.). We do not exhaustively explore settings nor compare with language models outside the GPT family. Experimentation leads to the following observations.

\begin{table}[htbp]
\centering
\footnotesize
\begin{tabular}{llcccc}
\toprule
\textbf{Model}  & \textbf{Input} & \multicolumn{4}{c}{\textbf{Metric}}\\
{} & {} & \scriptsize{purity} & \scriptsize{ARI} & \scriptsize{VM} & \scriptsize{purity$_{\textit{event}}$}  \\ 
\toprule

{embedding} & {text} & {0.96} & {0.95} & {0.95} & {4.36}  \\

{TF-IDF} & {text} & \textbf{0.98} & \textbf{0.97} & \textbf{0.97} & \textbf{5.47}  \\

{FEATHER} & {graph} & {0.82} & {0.20} & {0.33} & {4.09}  \\

{GAT} & {graph} & {0.83} & {0.46}  & {0.49} & {4.69}  \\

\bottomrule
\end{tabular}
\caption{\label{tab:results-clustering}
Results for causal graph clustering (higher is better). Evaluation is based on single labels (for first three metrics), with purity$_{\textit{event}}$ (Eq. \ref{eq:purity}) based on most frequent event types.}
\end{table}

\begin{table}[ht]
\centering
\footnotesize
\begin{tabular}{clcc}
\toprule
\textbf{Method}  & \textbf{Matching} &  \multicolumn{2}{c}{\textbf{Metric}}\\
{} & {} & {MAP}  & {MRR}  \\ 
\toprule

TF-IDF  & {text-to-text} & {0.36}  & {0.32} \\
GAT & {graph-to-graph}  & {0.48}  & {0.36} \\

\midrule

TF-IDF & {text-to-schema} & \uddot{0.59}  & {0.35} \\
GAT & {graph-to-schema}  & \uddot{\textbf{0.68}}  & \textbf{0.43} \\

\bottomrule
\end{tabular}
\caption{\label{tab:results-matching}
First (top), we match \textsc{Torquestra} to Wikipedia texts (MAVEN) using TF-IDF and graphs encoded with a graph attention network (GAT). For schema matching (bottom), we match MAVEN graphs to our causal schema library.}
\end{table}

\textbf{Large language models can generate complex structured representations}. Experiments show (Table \ref{tab:results}) we can generate symbolic causal knowledge in the form of directed, branching causal graph structures with multiple events and participants. We expect that research into interpretable, neuro-symbolic, stepwise reasoning using generative models may build upon this progress in structure prediction.

\textbf{Conditioning on structural knowledge improves generation performance, in some cases}. We evaluate if conditioning on temporal, event, and event type networks helps improve generated causal graph correctness and completeness and find that temporal and event structures appended to raw text result in more correct (Table \ref{tab:results}, \underline{+7\%}), more complete (\underline{\underline{+7\%}}) graphs than raw text alone. Overall, semantic signals jointly increase performance over conditioning on raw text alone using validation data (texts 100-150 tokens in length).

More specifically, we experimented with various forms of concatenated text and structures, including text alone, text + G$_{\textit{temp}}$, and text + G$_{\textit{temp}}$ + V$_{\textit{event}}$. With in-distribution data (top half of Table \ref{tab:results}), the most rich input (G$_{\textit{temp}}$ + V$_{\textit{event}}$) led to the best generation results. In contrast, with out-of-domain texts (bottom half), text alone works better than conditioning with `dense' paraphrases of V$_{event}$, with the length of test texts (2-4x longer than avg. validation) likely a factor.

\textbf{The student surpasses the teacher}. Following work in knowledge distillation \citep{West2022}, experiments show GPT2-XL$_{\textit{distill}}$ (student model, trained on knowledge graph triples, fine-tuned on \textsc{Torquestra}) outperforms few-shot GPT-3 (teacher model, trained to predict next word) (Table \ref{tab:results}, \udensdot{+10\%}). Further, GPT2-XL$_{\textit{distill}}$ outperforms original GPT2-XL in correctness (\udot{+8\%}) and completeness (\dashdot{+15\%}) (input text + G$_{\textit{temp}}$), evidence the distilled model learns causal structure, suggesting that we need commonsense knowledge for more complete and correct causal graph generation.

\textbf{Lexical methods for clustering texts are generally better}. Unsurprisingly, tf-idf significantly outperforms graph similarity methods across all metrics (Table \ref{tab:results-clustering}). We note tf-idf clusters are `quite' homogeneous, due in part to the provenance of the test data: Wikipedia articles automatically selected and labeled with topics \citep{Wang2020},  likely with similar methods as ours.

\textbf{Advantages of matching using graph methods versus words alone}. We measure similarity of event sequences for clustering comparing text-to-texts and graph-to-graphs, and for matching experiments comparing graph-to-schemas and text-to-schemas. Advantages of our system are evident matching graph-to-schemas (Table \ref{tab:results-matching}, \uddot{+9\%}). 

We find graph-based methods help identify articles with similar causal stories, e.g., graphs with a 4-nary node `military operation'. Graph-based methods provide a means of measuring conceptual similarities between the causal stories associated with events that may not otherwise be matched. For example, our algorithm finds a high similarity between the 1939 `Invasion of Poland' and a head-on train collision, where both stories involve opposing forces running into each other explosively, with similar (predictable) tragic consequences.

\textbf{Smaller block size helps identify salient subgraphs}. In training, setting block size (the length of input presented to the model) to shorter lengths (e.g., < 300 subword units) provides the model with only a subset of causal triples for each text. As we topologically sort input graphs using breadth-first search, the model learns to generate salient and connected edges. We leave for future work more rigorous evaluation of salience detection  using manual annotations we include as part of our data release.

\textbf{Evaluation is challenging}. There are many challenges associated with the evaluation of schema induction systems. On the one hand, lexical overlap and shared entities make two texts similar. On the other, similar causal structures, i.e. the causal schemas that stories share, can be discovered and compared. Still, the weighing of multiple semantic signals remains subjective.

We experimented with various means of evaluation: precision of topic labels (e.g., \textit{man-made disaster}), overlap of event types (text as bag-of-events) and subsets of frequent event types (Eq. \ref{eq:purity}). We qualitatively assess graph structural similarity, with an automated tool a work in-progress. 

We find that multiple measures of schema similarity to be more robust than using a single method, though we also recognize that more work, both theoretical and computational, needs to be done to develop still more reliable tools.

\textbf{Schema meaning}. Previous work views schemas as linear event orderings \citep{Chambers2013} and as more complex graph structures \cite{Li2021, Du2022}. How to further compose atomic meanings into larger semantic units for computational processing remains an open research question. Something like an event ontology of hierarchical event structures likely plays a role in the human conceptualization of event similarity, however, we make no hypotheses about better representations for computational applications.

\section{Conclusion}

We present \textsc{Torquestra}, a dataset of paired semantic graphs for studies of causal structure at the discourse level. Our experiments in causal graph generation, clustering, and schema matching provide insight into how to leverage \textsc{Torquestra} for knowledge discovery of latent causal structures of news texts, comparable to or outperforming search methods based on lexical similarity alone.

Research in knowledge discovery using causal schema induction will be of interest to historians, journalists, and health researchers looking for new angles on the study of narratives and stories. To support such research, we make our dataset, starter code, and evaluation tools publicly available\footnote{\url{https://fd-semantics.github.io/}}.

\section*{Acknowledgements}

Our special thanks to the annotation team at the University of Colorado Boulder for help in collecting data. Thanks to Ed Hovy, Yejin Choi, Dan Roth, Martha Palmer, Frank Ferraro, the XPO ontology group, and Heng Ji for guidance, inspiration, and feedback. This research was supported in part by DARPA under I2O (RA-21-02) and DARPA under the KAIROS program (FA8750-19-2-1004).

\bibliography{anthology}
\bibliographystyle{acl_natbib}

\newpage

\appendix
\input{appendix}

\end{document}

%% file: appendix.tex
\onecolumn

\section{Appendix}
\label{sec:appendix}

\subsection{Limitations}
\label{ssec:limitations}

\textbf{Limited scales of causation}. We rely on a limited number and diversity of viewpoints for scales of causation (eight annotators at undergraduate and graduate levels at U.S. institutions of higher learning; four female, four male; five native, three non-native English speakers).

\textbf{Size of dataset}. In Table \ref{tab:datasets}, we see that \textsc{Torquestra} is slightly larger than other existing human-curated temporal and event structure reasoning datasets. Still, the question remains, how large must the corpus be to be able to enable successful learning of salient discourse-level explicit and implicit causal relations? It might be significant.  

\textbf{More and more varied test data}. A collection of 3k Wikipedia articles in a single language is a relatively small sample for testing. We note the document label class imbalance: about 33\% concern historical military conflicts. Also, labeled event types are noisy (e.g., ``The area was \underline{hit} by a storm'' is labeled \textsc{Violence-Attack}). Topic labels were likely originally assigned using similar methods as our own (tf-idf).

\textbf{Schema indeterminacy}. For better schema matching models, it is possibly although not necessarily imperative to define in clear terms what a ``schema'' is. A ``better'' schema model is likely hierarchical in terms of event semantics, although the compositional nature of events is still indeterminate, and thus also generic events (schemas). Different manners of defining schemas are possible, some of which we explore in this work: as single event types, as sequences of event types (e.g., in a causal chain), as unordered sets of event types, as causal graphs, as having predicate argument structure, etc.

\textbf{Causal relations and other possible experiments}. We note that we have not reported on experiments with the causal subrelations shown in Table \ref{tab:relations}, work we leave for the future. We also have not reported on experiments ranking causal chain salience, nor on generation results directly conditioning on $S_{\textit{causal}}$ (limiting the use of causal schema graphs to matching experiments), a promising research direction. Data for each of these proposed experiments is available as part of our release.

 \textbf{Role of structured representations in NLP}. The role of structured representations in NLP is  debatable in the deep learning era (the importance of dependency parsing, syntactic treebanks, semantic role labeling, etc.). One concern is that such human annotation is expensive. Another concern is that in many cases linguistic structure can be preempted, using raw text only for both inputs and outputs.  
 
 To add to this debate, the success of methods in causal inference is largely based on the use of graphical structures, though there has not been yet a direct link made between how to automatically acquire these structures from language text using NLP methods, a concern we hope the present work helps address.

\pagebreak

\subsection{Describing Torquestra}
\label{ssec:describing-torquestra}

\begin{figure*}[htbp]
\begin{minipage}{.48\linewidth}
\captionsetup{}
  {\includegraphics[width=215pt]{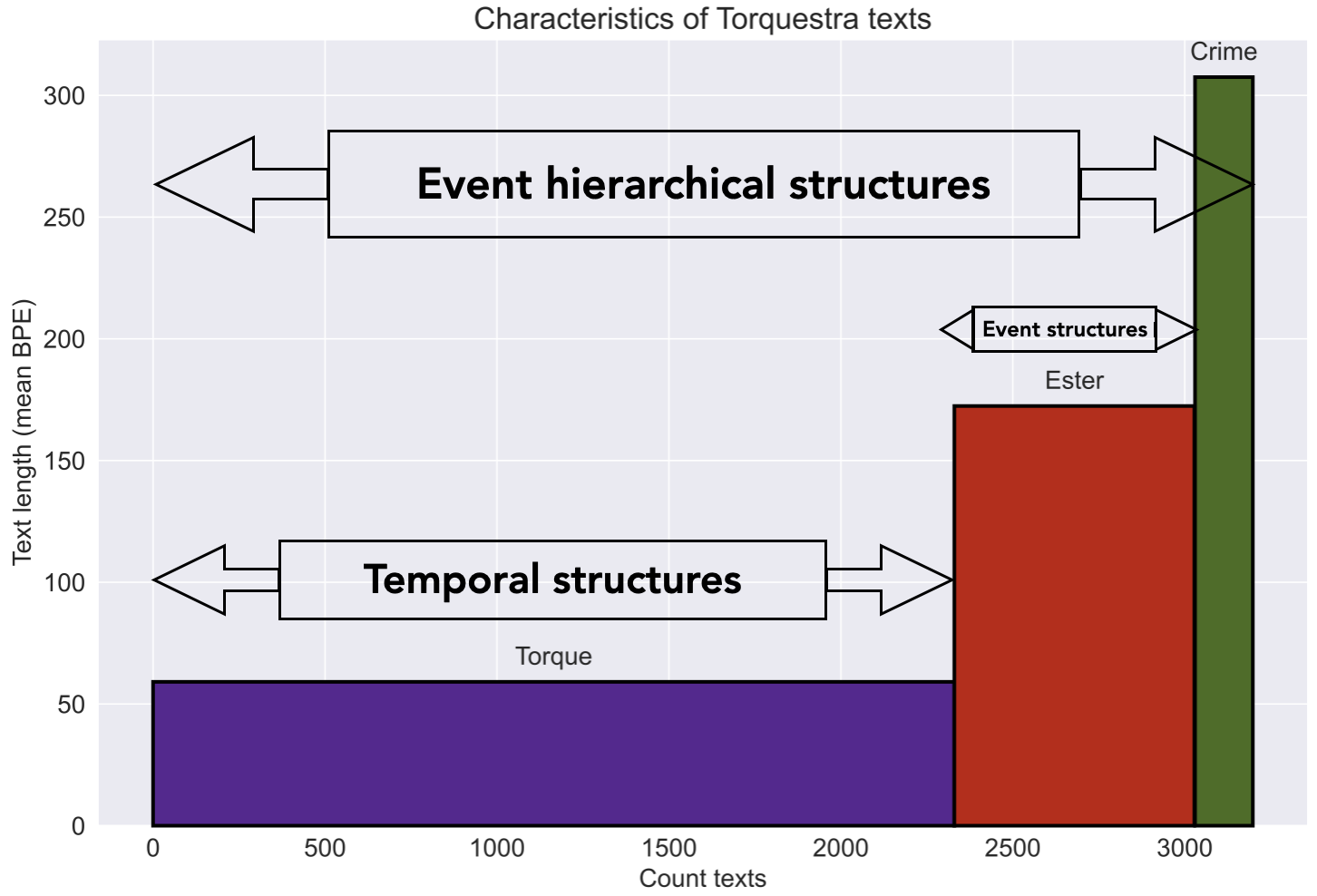}}
  \caption{\label{fig:torquestra-content}Count of texts for different slices of \textsc{Torquestra} by text mean length  (\#subword tokens). Existing human-engineered temporal and event structures  (in boxes with arrows) supplement our causal structures (see Fig. \ref{fig:torquestra-graphs} on right).}
\end{minipage}
\hspace{0.02\linewidth}
\begin{minipage}{.48\linewidth}
\centering
\captionsetup{}
 {\includegraphics[width=195pt,trim={1.2cm 0.6cm 1.2cm 1.2cm},clip]{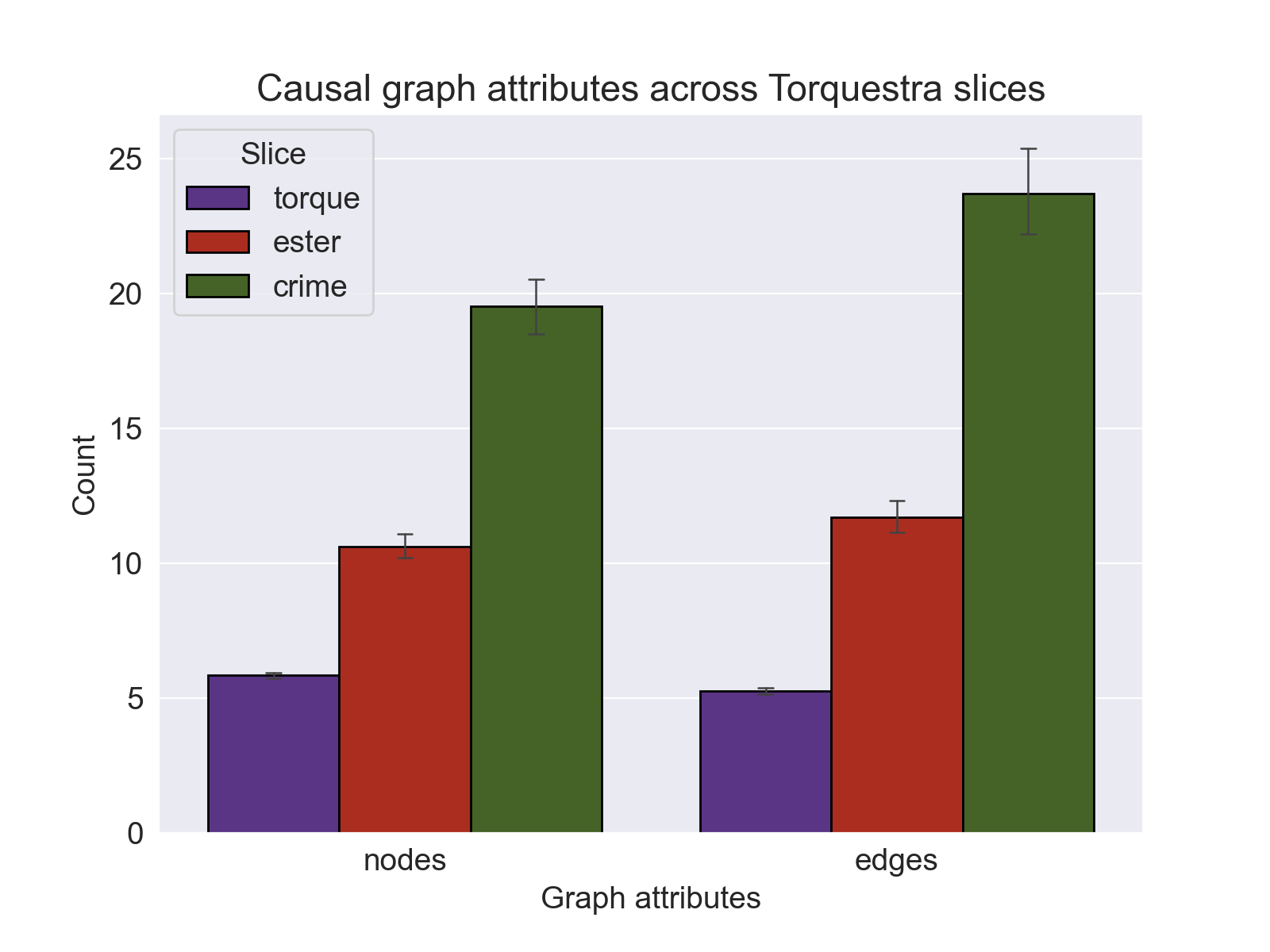}}
  \caption{\label{fig:torquestra-graphs}Complexity of human-engineered causal graphs in \textsc{Torquestra}. The \textsc{Torque} data slice (purple, leftmost) consists of  less complex causal graphs (based on number of nodes and edges) compared to \textsc{Crime} (green, rightmost).}
\end{minipage}
\end{figure*}

\textbf{Text and graph statistics}

\begin{table*}[htbp]
\centering
\small
\begin{tabular}{llllll}
\toprule
\scriptsize\textbf{Data attribute} & \scriptsize{$G_{\textit{temp-madaan}}$} & \scriptsize{$G_{\textit{temp-torque}}$} & \scriptsize{$G_{\textit{causal-torque}}$} & \scriptsize{$G_{\textit{causal-ester}}$} & \scriptsize{$G_{\textit{causal-crime}}$} \\
\midrule
\small
avg \# tokens/text &	{120.3} & {46.8}    &	{46.8}  &	 {133.4}  &	{231.3}   \\
\midrule
max \# nodes &	{} & {22.0}    &	{17.0}  &	   {19.0}  &	{36.0}   \\
mean \# nodes & {3.3} & {8.3}   &	{5.81} &	{11.6}  &	{19.5}   \\
std \# nodes &	{} & {3.2} &	{2.3} &	{3.27}  &	{5.93}   \\
\midrule
max \# edges &	{} & {102.0} &	{21.0}      &	{24.0}  &	{48.0}  \\
mean \# edges & {3.9} & {20.9} &	{5.2}    &	{13.1}  &	{23.7}  \\
std \# edges &  {} & {13.4} &	{2.9}   &	{4.3}  &	{8.6} \\
\midrule
mean \# enables & {} & {} &	{4.3}      &	{11.5}  &	{21.1} \\
mean \# blocks 	& {} & {} &	{0.9}      &	{1.6}  &	{2.6}  \\
\midrule
mean degree     &  {}   & {} &   {1.7}  &	{2.3}  &	{2.4}   \\
mean clustering & {}    & {} &	{0.04}   &	{0.08}  &	{0.07}   \\
mean transitivity & {}  & {} &	{0.05}   &	{0.09}  &	{0.07}   \\
mean sq clustering & {} & {} &	{0.02}   &	{0.05}  &	{0.03}  \\
\bottomrule
\end{tabular}
\caption{\label{tab:torquestra-statistics}
Comparison of data statistics from two temporal structure resources: $G_{\textit{temp-madaan}}$ \citep{Madaan2021a} and $G_{\textit{temp-torque}}$ \citep{Ning2020} with various slices of \textsc{Torquestra}: $G_{\textit{causal-torque}}$, $G_{\textit{causal-ester}}$ and $G_{\textit{causal-crime}}$. One important comparison here is between the number of edges between $G_{\textit{temp}}$ (mean 20.9) versus $G_{\textit{causal-torque}}$ (mean 5.2), which we could use to argue that causal structures provide a cleaner starting point for reasoning than temporal structures.
}
\end{table*}

\pagebreak

\subsection{Analysis of events in \textsc{Torquestra}}
\label{ssec:appendix-events}

\begin{figure*}[h!]
\centering
\includegraphics[width=0.95\textwidth]{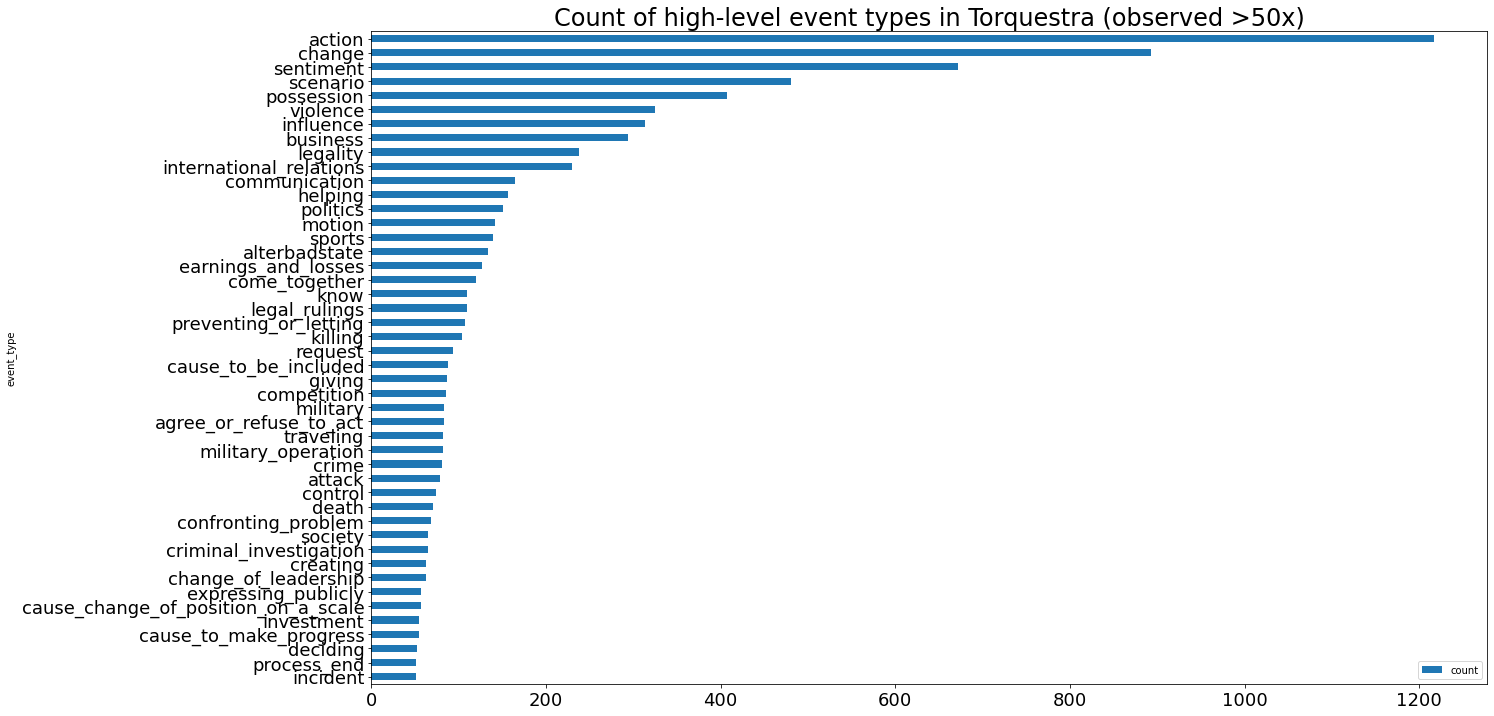}
\caption{\label{fig:high-level}Count of high-level hierarchical event semantic types observed in \textsc{Torquestra} texts.}
\end{figure*}

\begin{figure*}[h!]
\centering
\includegraphics[width=0.95\textwidth]{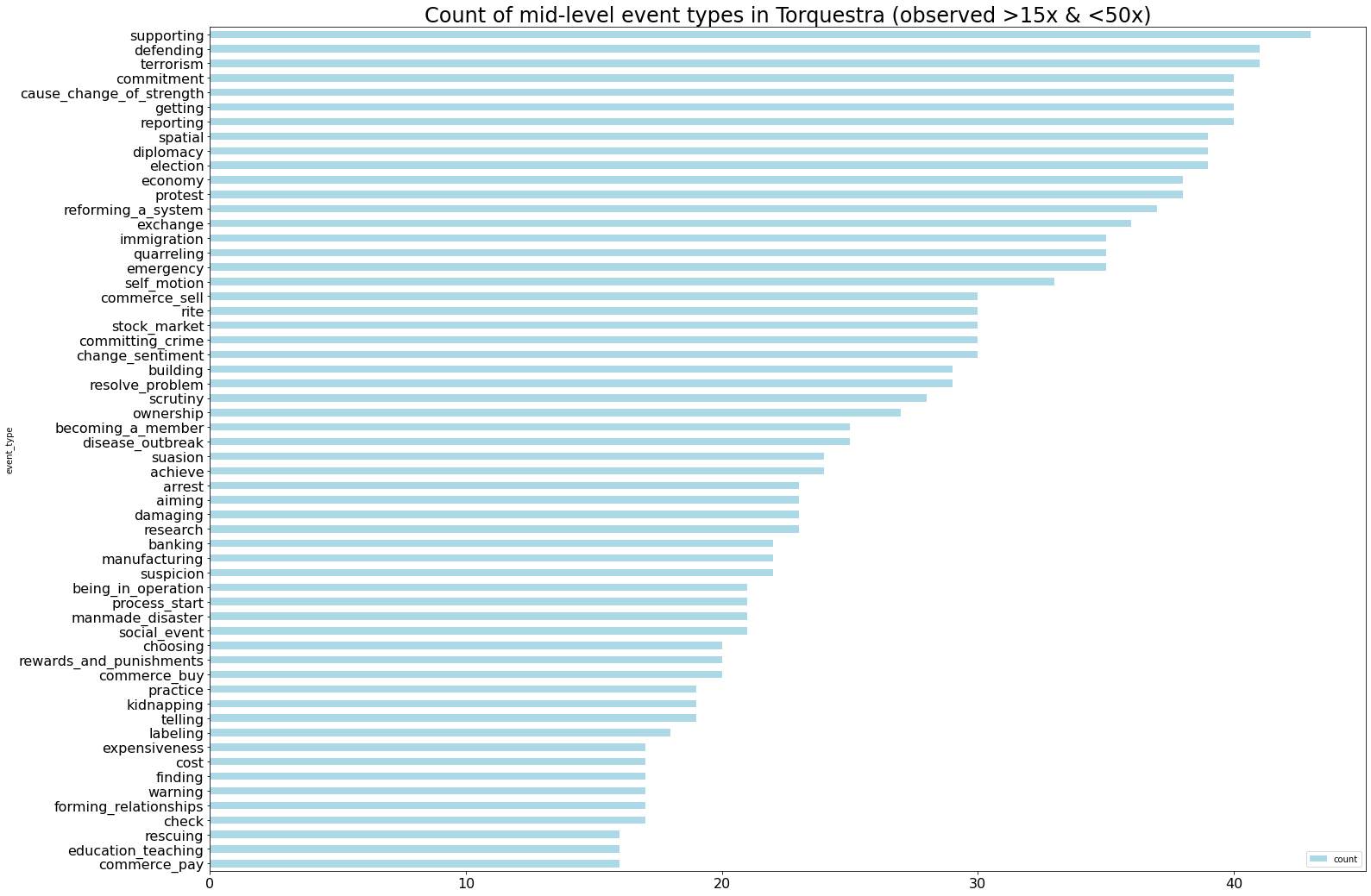}
\caption{\label{fig:mid-level}Count of mid-level hierarchical event semantic types observed in \textsc{Torquestra} texts.}
\end{figure*}

\pagebreak

\subsection{Annotation}
\label{ssec:appendix-annotation}

\textbf{General annotation guidelines}. We refer to our approach of causal modeling as `participant-centered causal structure' (\S\ref{sec:definitions}). In this approach, nodes in causal graphs can be events expressed as sentences or phrases of Subject-Verb-Object form (e.g., `Alice wrote a paper'). A range of other event descriptions are possible (regarding e.g., nominalizations, modality, polarity). Nodes can also be event participants such as people and objects that directly contribute to, initiate, or disrupt the beginning, unfolding, or ending of events. `Alice made a cake' may be thus `Alice' (as causal agent) $\rightarrow$ `Alice made a cake'.

Annotators were asked to keep clearly in mind that one is not trying to model true causation, but just the apparent causation within the belief system of the speaker. Thus, causal links roughly mean ``in the opinion of the speaker there should be a causal relation between A and B'' — whether or not the actual causal relation (Table \ref{tab:relations}) is explicitly stated or implied somehow. This guideline helps remove the problem of reasoning about the world and allows one to frame debugging questions in terms of what the speaker might say and do, things that are much easier to discuss and evaluate. 

\textbf{Annotation quality}. Consistent labeling of graph nodes proved challenging. For free-form node labeling, annotators were asked to compose a short sentence or phrase of form \textit{Subject-Verb-Object}, e.g. `Police arrested the suspect', or `suspect arrested'. For evaluation, we selected 50 texts, asked two annotators to write short texts for nodes, counting the number that shared at least one content word. For the 325 nodes evaluated, 56\% shared 1+ tokens referring to an event or other participant.

\textbf{Notes about salient causal chain identification}. Event salience detection is often framed as identifying the single most reportable event in an event sequence \citep{Ouyang2015}. We extend this task to be: Given a set of triples, identify a subset of triples that tell the central part of the causal story. For evaluation, dual annotations for 50 causal graphs show agreement of $\kappa=0.61$.

\begin{figure}[h!]
\centering\includegraphics[width=0.45\textwidth]{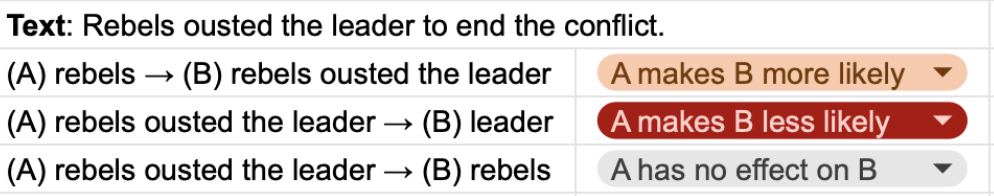}
    \caption{To evaluate a causal relation given a text and triple (rows), we used intuitive causal notions such `A makes B more likely' in place of directly assessing an \textsc{Enables} edge.}
    \label{ex:rebels-eval}
\end{figure}

\textbf{Notes about data evaluation: causal relations}. For evaluation of human and machine generated causal graphs, annotators are given a text, two participants (\textbf{A} and \textbf{B}) in a directed relation, and three choices: A makes B more likely (\textsc{Enables}), less likely (\textsc{Blocks}) or has no effect, see Fig. \ref{ex:rebels-eval}. For 150 edges judged by three evaluators, average pairwise Cohen's kappa is $\kappa=0.53$ (moderate agreement). 

\subsection{Notes about data evaluation: causal graph completeness}

We concede that causal graphs are difficult to fully specify for various reasons. First, causal expectations do not hold in certain contexts, implicit contributory factors related to events are typically beyond enumeration, and causal relations for events with negative polarity are easy to overlook.

\textbf{Causal expectations} for events that fail in a given context are difficult to agree on, e.g., subrelations \textsc{Without effect} and \textsc{Unknown}, $\approx$15\% of relations in \textsc{Torquestra}. In more than half of observed cases, explicit lexical cues (e.g., ``despite'' in ``The protest happened despite the rain.'') are not present.

\textbf{Implicit contributory factors} are numerous in certain kinds of text. In our case, newswire is necessarily succinct leaving a lot unsaid. Commonsense often suggests causal nodes of the kind ``and etc.,'' perhaps shorthand for ``there are many other potential contributory factors that I cannot all list, but they fit here.'' We leave the collection of this type of data for future work. 

\textbf{Notes about event polarity}. Negative event polarity is a challenge for representation. There may be a reporting bias, i.e. perhaps people less frequently speak about the causal effects of negative polarity events, e.g., `It didn't rain, so I didn't play in the mud'. Some of these semantics are captured in the \textsc{Without effects} causal subrelation. Annotators were instructed to look out for causal relations not covered by our ontology and to construct nodes including negative polarity, when possible. Additional thought and corpus studies may point to additional layers of meaning distinctions possible.

\subsection{Prompt engineering}
\label{ssec:prompts}

We construct various prompts for conditional language model generation. Using the pre-existing temporal and event structure annotations, prompts can be made concatenating raw texts with associated questions and answers about temporal and event structures. We also experiment by adding knowledge of hierarchical event types directly into texts, using dense paraphrasing \citep{Tu2022}.

\textbf{Example prompt: text + G$_{\textit{temp}}$}. For prompts for text + G$_{temp}$, we randomly selected 2-3 temporal \textsc{Torque} questions and answers \citep{Ning2020} in the form of lists of event mentions, filtering out verbs with less eventive meaning (e.g., `was') and appended to the source text.

\begin{small}
\begin{verbatim}
TEXT: And on that basis Keating was released from prison before he was eligible for parole. 
Now the ninth US circuit court of appeals has ruled that the original appeal was flawed 
since it brought up issues that had not been raised before.\n
What event has already finished? released, ruled, brought, flawed\n
What happened before the court ruled? flawed, brought, raised, released\n
What did not happen before Keating was released? parole\n
[GEN]
\end{verbatim}
\end{small}

\textbf{Example prompt: text + V$_{\textit{event}}$} (cf. dense paraphrasing). We also add structure to input text prepending event types to event and entity mentions. We wonder: Does it \textit{help} or \textit{hinder} generation by: 1) indicating which tokens are likely associated with causal chains? and, 2) giving an explicit signal of hierarchical event semantics? We also include event types for entities and events NOT part of causal chains (to aid in learning). Results are split on this question: in some cases, it helps, in other cases, it hinders (\S\ref{sec:results}).

\begin{small}
\begin{verbatim}
TEXT: And on that basis Entity::Keating was Legal_rulings::released from Entity::prison 
before he was Scenario::eligible for parole. 
Now the Entity::ninth US circuit court of appeals has Legal_rulings::ruled that 
the original Legal_rulings::appeal was Incident::flawed 
since it brought up Scenario::issues that had not been raised before.\n
[GEN]
\end{verbatim}
\end{small}

\subsection{Model specifics}

We implemented graph generation models in PyTorch using the huggingface library. Without a limited hyperparameter search, we note generation models worked well with the following: block size between 300-600, training 12-16 epochs, learning rate=$3e^{-5}$, cosine learning rate scheduler, and Adam optimizer. For generation, we use nucleus sampling \citep{Holtzman2019} setting $p=0.90$ and temperature$=0.95$. 

To embed causal graphs, we implement a self-supervised Graph Attention Network (GAT) \citep{Velickovic2018} using PyTorch Geometric\footnote{\url{https://pytorch-geometric.readthedocs.io/en/latest/}}. Our base model uses two GAT layers with ELU activations, dropout=$0.5$, global mean pooling and a final linear layer. We trained for 2-3 epochs over the entire dataset (\textsc{Torquestra} + test data). Using the trained model, we make final embeddings for clustering experiments (test set alone); while for matching we compare the test set with \textsc{Torquestra} graphs and RESIN subgraphs.

\textbf{Masked objective function}. For the self-supervised graph modeling, we train a graph neural network using message passing to compose graph embeddings, randomly masking input nodes. In training, we hypothesize that more important nodes will contribute more to a final graph embedding, and experiment  masking nodes based on PageRank weighting\footnote{Using the NetworkX Python package, \url{https://networkx.org/}}.

\textbf{Loss function}. At each step of training, the model `sees' 5 of 10 randomly sampled masked graphs, and predicts a graph embedding. We compare the similarity of the predicted embedding for the masked graph, $y*$, with the embedding for the entire ground truth graph, $y$ and train the model to minimize an absolute value cosine similarity difference loss:
\begin{equation}
\label{eq:loss}
\mathcal{L}_{CosineSimDiff} = |1 - CosineSimilarity(y*, y)|
\end{equation}
\noindent
Cosine similarity, the scaled dot product of non-zero vectors \textbf{A} and \textbf{B}, is:

\begin{equation}
CosineSimilarity(\textbf{A}, \textbf{B}) = \frac{\textbf{A}\cdot\textbf{B}}{max(\norm{\textbf{A}}_2\norm{\textbf{B}}_2, \epsilon)} 
\end{equation}
\noindent
with $\epsilon=1e^{-8}$ to prevent division by zero. For vectors that are equal, the cosine similarity is 1, 0 when orthogonal, and -1 when opposite, so the absolute value cosine similarity difference loss (Eq. \ref{eq:loss}) approaches zero as the embedding similarity for masked graph and non-masked graph increases. In our experiments, we mostly observed loss to decrease from 1 to about 0.6 in training and validation.

\raggedbottom
\subsection{Metrics}
\label{ssec:metrics}

For our experiments, we report a mixture of quantitative and qualitative metrics based on 2-3 human judgments. However, ground truth labels for nodes and schemas present a number of tricky issues (see Appendix \ref{ssec:limitations}). This is apparent, for example, in judgments of generated graph correctness and also in the evaluation of schema matching models. Defining schemas as a single label is likely too simple, a situation that using a set of labels does not necessarily improve.

For schema matching, we report mean average precision (MAP), measured between 0 and 1 with higher being better. For Q queries, this is defined as:
\begin{equation}
MAP = \frac{1}{Q} \sum_{q=1}^{Q} \text{avg.  precision}(q)
\end{equation}

with precision for a query $q$ defined as:
\begin{equation}
\text{avg. precision}(q) = \frac{\abs{\{\text{relevant documents}\}_q\cap{\{\text{retrieved documents}\}_q}}} {\abs{\{\text{retrieved documents}\}_q}}
\end{equation}

We consider different quantitative and qualitative judgments for relevance, including: precision of topic label, lexical overlap and structural similarity, where structural similarity can be defined in terms of shared nodes, node degree, n-nary properties, and other observable graph properties. See a cluster example sharing structural similarity in Appendix \ref{ssec:cluster}.

\pagebreak
\subsection{Data sample}

Training examples are shorter texts from TORQUE (with temporal questions); Validation are longer texts from ESTER (with event structure questions) that \textbf{includes} the shorter TORQUE text, in \textbf{bold}.

\begin{scriptsize}
\begin{lstlisting}[escapechar=\%]
{`split': `train', 
`source': `torque', 
`@id': `train-docid_PRI19980115.2000.0186_sentid_6',
`notes': `original-698', 
`text': `%\textbf{And on that basis Keating was released from prison before he was eligible for parole. Now the ninth US circuit court of appeals has ruled \\ \indent that the original appeal was flawed since it brought up issues that had not been raised before.}%', 
`questions': 
    [`What event has already finished?',
    `What event has begun but has not finished?',  
    `What will happen in the future?', 
    `What happened after Keating was released?', 
    `What did not happen before Keating was released?', 
    `What happened before the court ruled?', 
    `What did not happen after Keating was released?', 
    `What happened after the court ruled?'], 
`answers': 
    [[`released', `ruled', `brought', `flawed'], [], [], 
    [`ruled', `flawed', `brought', `raised'], 
    [`was', `parole'], 
    [`flawed', `brought', `raised', `released'], [], []], 
`event_types': 
    {`Keating was released': `Action;Legality;Legal_rulings;Releasing', 
    `Keating was eligible for parole': `Action;Legality;Legal_rulings', 
    `issues that had not been raised before': `Scenario', 
    `court ruled that the original appeal was flawed': `Action;Legality;Legal_rulings'}, 
`noncausal_event_types': 
    {`brought': 'Action;Communication;Reporting'}, 
`causal_graph': [
    {`head': 'court ruled that the original appeal was flawed', `rel': `ENABLES', 
    `tail': 'Keating was released'}, 
    {`head': 'Keating was eligible for parole', `rel': 'BLOCKS', 
    `tail': 'Keating was released'}, 
    {`head': 'issues that had not been raised before', `rel': `ENABLES', 
    `tail': 'ruled that the original appeal was flawed'}]}

{`split': `dev', 
`source': `ester', 
`@id': `dev-docid_PRI19980115.2000.0186_sentid_6', 
`notes': `original-698', 
`text': "Former savings and loan chief, Charles Keating, is facing more legal troubles in California. 
    A federal appeals court has reinstated his state convictions for securities fraud. 
    NPR's Elaine Corey has more from San Francisco. In nineteen ninety-one Charles Keating was convicted 
    in state court of helping to defraud thousands of investors who bought high risk junk bonds 
    sold by Keating's employees at Lincoln savings and loan. The bonds became worthless when the bankrupt thrift 
    was seized by government regulators. Keating's convictions were thrown out in nineteen ninety-six on a technicality. 
    %\textbf{And on that basis Keating was released from prison before he was eligible for parole. Now the ninth US circuit court of appeals has ruled \\ \indent that the original appeal was flawed since it brought up issues that had not been raised before.}%
    That means the convictions stand, a ruling likely to send Keating's lawyers 
    back to state court where they must start over with a new appeal.", 
`questions': 
    [`Why was Mr. Keating convicted?', `Why was Mr. Keating released from prison?', 
    `What might happen as a result of the convictions being ruled to stand after the flawed appeal?'], 
`answers':
    [`defraud thousands of investors', 
    `convictions were thrown out in nineteen ninety-six on a technicality', 
    "send Keating's lawyers back to state court"], 
`event_types': `banking;crime;action;legality;legal_rulings', 
`causal_graph': 
    [{`head': `Entity::Charles Keating', `rel': `ENABLES', 
    `tail': `Keating faces legal troubles', `saliency': 0}, 
    {`head': `Entity::Charles Keating', `rel': `ENABLES', 
    `tail': `security fraud', `saliency': 0}, 
    {`head': `security fraud', `rel': `ENABLES', 
    `tail': `Keating is convicted of security fraud', `saliency': 1}, 
    {`head': "Keating's convictions dismissed", `rel': 'BLOCKS', 
    `tail': `Keating is convicted of security fraud', `saliency': 1}, 
    {`head': "Keating's convictions dismissed", `rel': `ENABLES',
    `tail': `Keating was released from prison', `saliency': 1}, 
    {`head': `technicality', `rel': `ENABLES', 
    `tail': "Keating's convictions dismissed", `saliency': 0}, 
    {`head': `bonds became worthless', `rel': `ENABLES', 
    `tail': `security fraud', `saliency': 0}, 
    {`head': `bankrupt thrift was seized', `rel': `ENABLES', 
    `tail': `bonds became worthless', `saliency': 0}, 
    {`head': `Entity:ninth US circuit court of appeals', `rel': `ENABLES', 
    `tail': "court reinstated Keating's state convictions for securities fraud", `saliency': 1}, 
    {`head': "court reinstated Keating's state convictions for securities fraud", `rel': `ENABLES', 
    `tail': `Keating faces legal troubles', `saliency': 1}, 
    {`head': "Keating's convictions dismissed", `rel': `ENABLES', 
    `tail': "court reinstated Keating's state convictions for securities fraud",  `saliency': 1}]}
\end{lstlisting}
\end{scriptsize}

\pagebreak

\section{Cluster example}
\label{ssec:cluster}

\begin{figure*}[h]
\centering
\includegraphics[width=0.70\textwidth]{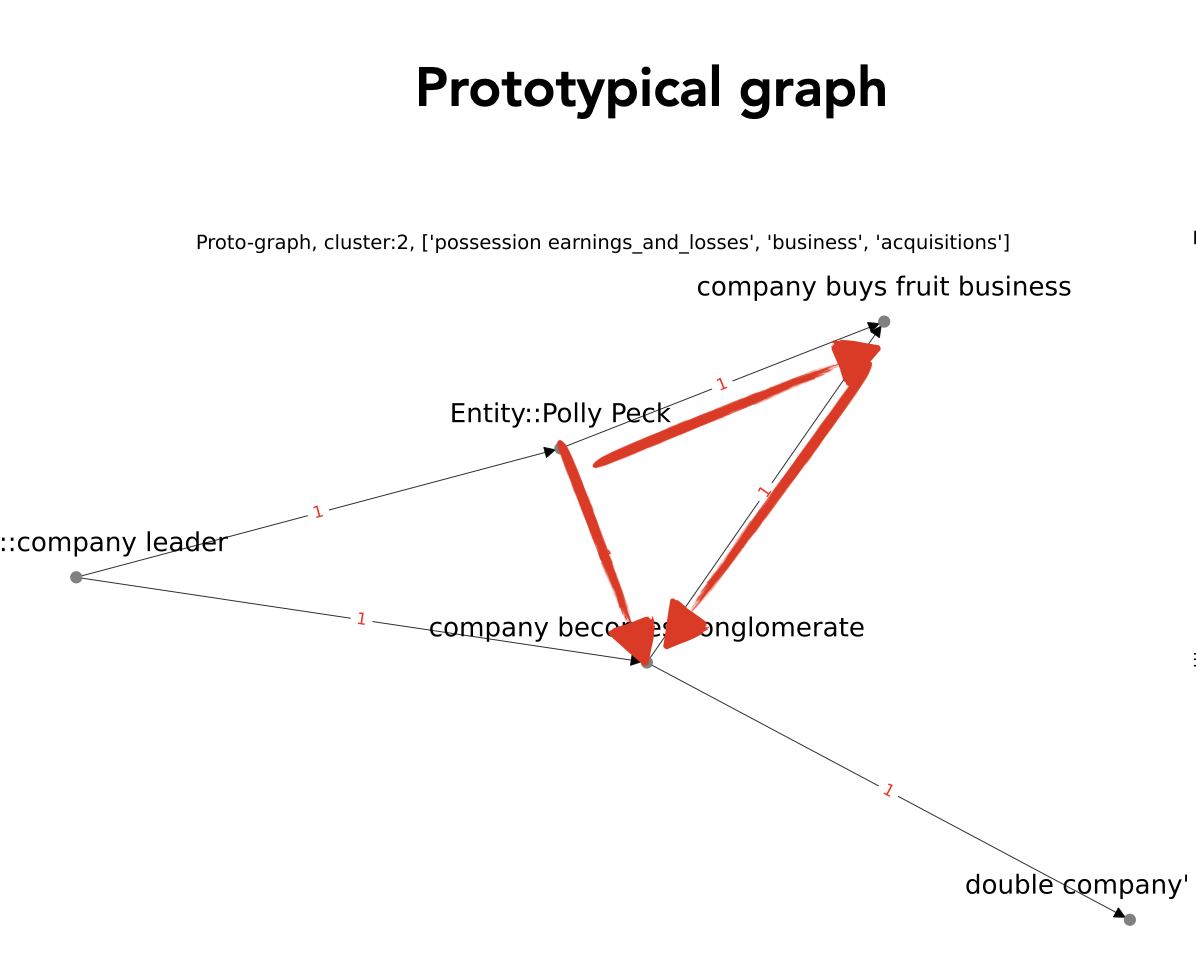}
\caption{\label{fig:cluster1} Example prototypical graph (cluster centroid) in an early experiment clustering \textsc{Torquestra}$_{\textit{human}}$ causal graphs.}
\end{figure*}

\begin{figure*}[h!]
\centering
\includegraphics[width=0.80\textwidth]{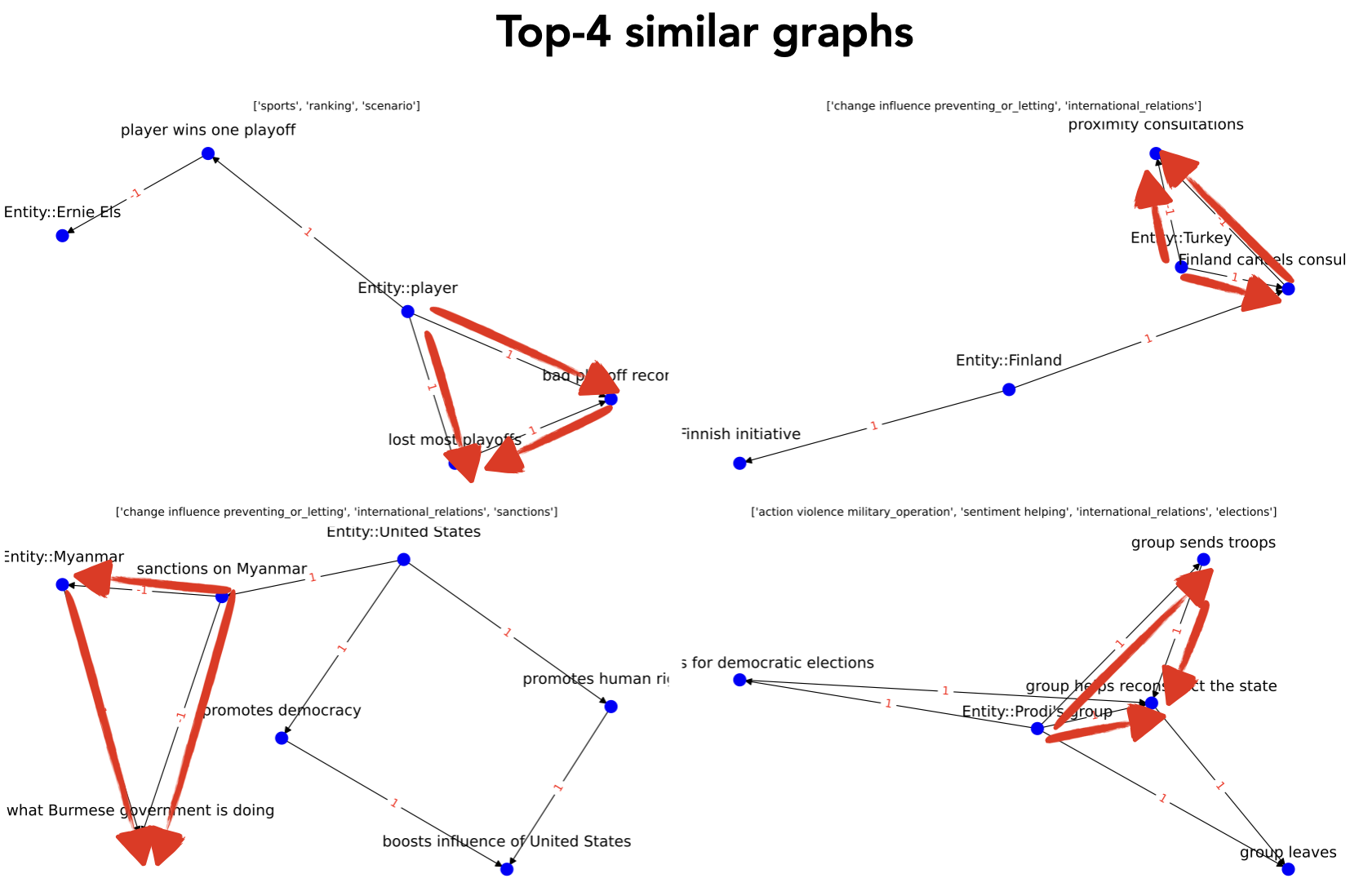}
\caption{\label{fig:cluster2}Similar graphs from a cluster. The GNN matching algorithm is effective in identifying similar graph structures as well as those that share lexico-semantic similarity, cf. \textit{motif discovery} \citep{Milo2002}.}
\end{figure*}